\documentclass[table]{article}

\usepackage{arxiv}

\usepackage{amsmath}

\usepackage[utf8]{inputenc} 
\usepackage[T1]{fontenc}    
\usepackage{hyperref}       
\usepackage{url}            
\usepackage{booktabs}       
\usepackage{amsfonts}       
\usepackage{nicefrac}       
\usepackage{microtype}      
\usepackage{cleveref}       
\usepackage{graphicx}
\usepackage[numbers, square]{natbib}
\usepackage{doi}

\usepackage{tikz}
\usetikzlibrary{fadings}
\usepackage{tcolorbox}
\usepackage{mathtools}
\usepackage{amsthm}
\usepackage{bbm}
\usepackage{bm}
\usepackage{tabularx}
\usepackage{threeparttable}
\usepackage{makecell}
\usepackage{multirow}
\usepackage{booktabs}
\usepackage{subcaption}
\usepackage{flushend}
\usepackage{dirtytalk}
\usepackage{xcolor}
\usepackage{circuitikz}
\usepackage{amsmath}
\usepackage{algorithm}
\usepackage{algorithmic}
\usepackage{enumitem}
\usepackage{hyperref}

\title{On the Impact of Anonymization on the Performance of Large Language Models}

\date{}

\usepackage{authblk}

\author[$\dagger$,1,2,3]{%
	Tobias Deu{\ss}er\thanks{\texttt{tdeusser@uni-bonn.de}, ORCID-ID: 0000-0003-4685-0847}%
}
\author[$\dagger$,1,2]{%
	Max Hahnb{\"u}ck
}
\author[1,3]{%
	Lorenz Sparrenberg
}
\author[4]{%
	Tobias Uelwer
}
\author[1,2,3]{%
	Christian Bauckhage
}
\author[1,2,3]{%
	Rafet Sifa
}
\affil[$\dagger$]{\emph{These authors contributed equally to this work.}}
\affil[1]{University of Bonn, Bonn, Germany}
\affil[2]{Fraunhofer IAIS, Sankt Augustin, Germany}
\affil[3]{Lamarr Institute for Machine Learning and Artificial Intelligence, Bonn, Germany}
\affil[4]{Microsoft Germany GmbH, Cologne, Germany}

\renewcommand{\headeright}{}
\renewcommand{\undertitle}{}
\renewcommand{\shorttitle}{On the Impact of Anonymization on the Performance of Large Language Models}

\hypersetup{
pdftitle={On the Impact of Anonymization on the Performance of Large Language Models},
pdfsubject={cs.CL, cs.AI},
pdfauthor={Tobias Deu{\ss}er, Max Hahnb{\"u}ck, Lorenz Sparrenberg, Tobias Uelwer, Christian Bauckhage, Rafet Sifa},
pdfkeywords={anonymization, large language models, natural language processing, machine learning},
}

\begin{document}
\maketitle
\begin{abstract}
As large language models are increasingly deployed in sensitive domains, anonymizing input data to protect personally identifiable information has become a critical practice. However, the impact of this anonymization on model utility is not well understood. This paper presents a systematic empirical study of the trade-off between privacy and performance. We evaluate five prominent language models across eleven diverse benchmarks, comparing their performance on original versus pseudonymized inputs. Our results reveal that while anonymization generally degrades performance, the effect is highly nuanced. We find that more capable models, such as Qwen2.5-72B and GPT-4o mini, suffer the largest performance drops, suggesting a stronger reliance on specific entity information. The impact is also task-dependent: performance on TruthfulQA improves with anonymization, while retrieval-focused tasks like RGB experience a catastrophic decline. Further experiments show that reversible anonymization techniques that preserve entity uniqueness significantly outperform irreversible ones like redaction, and that explicitly prompting models about anonymization offers no discernible benefit. We conclude that anonymization is not a one-size-fits-all solution and must be co-designed with the model and task in mind to balance privacy and utility effectively. Our findings provide a crucial baseline for developing more robust, privacy-aware AI systems.
\end{abstract}


\section{Introduction}


Large Language Models (LLMs) have become increasingly central to modern artificial intelligence applications, from analyzing medical reports \citep{doshi2024quantitative}, writing legal documents \citep{cui2024chatlaw}, finding contradictions in financial documents \citep{deusser2023uncovering}, connecting patients with rare diseases with each other \citep{berger2024optimizing}, to enabling automated auditing for financial reports \citep{hillebrand2023improving}. Yet, as such LLM approaches are deployed in real-world settings, privacy concerns are becoming more and more prevalent \citep{yao2024survey,das2025security}. One common mitigation strategy is anonymization, i.e., removing or masking personally identifiable information (PII) before feeding inputs into a model \citep{biesner2022anonymization,sepas2022algorithms,deusser-etal-2025-resource}. 

But what does this actually cost us in terms of performance? In this paper, we take a systematic look at the tradeoff between privacy and capability. We evaluate a set of five models (see Table~\ref{tab:anon_llm_overview}) spanning a wide range of sizes and capabilities across eleven diverse benchmarks (see Table~\ref{tab:anon_benchmark_overview}). These benchmarks cover a spectrum of natural language processing (NLP) tasks, including general scientific reasoning (ARC, \citealp{clark2018thinksolvedquestionanswering}), medical question answering (MedQA, \citealp{jin2021disease}), emotional intelligence (EQ-Bench, \citealp{paech2024eqbenchemotionalintelligencebenchmark}), and translation (WMT 2014, \citealp{bojar2014findings}), allowing us to assess the trade-off across different cognitive and linguistic domains.

We observe that, while anonymization generally degrades performance as expected, the magnitude of this effect is highly dependent on the model's capability, the nature of the task, and the specific anonymization technique employed. Our key findings are threefold:
\begin{enumerate}
    \item \textbf{Model capability is inversely correlated with robustness to anonymization.} More powerful models like Qwen2.5-72B \citep{qwen2025qwen25technicalreport} and GPT-4o mini \citep{OpenAIGPT4omini2024} experience the largest performance drops, suggesting they rely more heavily on specific entity information to achieve their state-of-the-art performance.
    \item \textbf{Task sensitivity varies dramatically.} Performance on retrieval-intensive benchmarks like RGB collapses catastrophically post-anonymization. Conversely, on TruthfulQA, a benchmark designed to test factual accuracy, performance consistently improves, suggesting that removing named entities can mitigate a model's tendency to hallucinate incorrect associations.
    \item \textbf{Technique and prompting matter, but not always as expected.} Reversible anonymization techniques like pseudonymization, which preserve entity uniqueness, significantly outperform irreversible methods like redaction. However, explicitly prompting a model that its input has been anonymized yields no discernible performance benefit.
\end{enumerate}

Our results challenge the notion of anonymization as a one-size-fits-all solution, demonstrating that it must be co-designed with the model and task in mind. These findings provide actionable insights for practitioners and a foundation for the discussion on and development of more robust, privacy-aware, and ultimately more trustworthy NLP systems.

\section{Related Work}
The need to anonymize text data to protect personally identifiable information (PII) is a well-established practice driven by regulatory frameworks and the goal of developing trustworthy AI systems \citep{biesner2022anonymization, sepas2022algorithms}. However, this introduces a fundamental tension between preserving privacy and maintaining data utility for downstream applications \citep{Lison2021anonymisation}. Our work contributes to the growing body of research addressing this privacy-utility trade-off, with a focus on the capabilities and shortcomings of modern large-scale language models (LLMs).

Early anonymization systems predominantly combined named entity recognition (NER), rule-based filters, and external gazetteers to mask explicit identifiers \citep{Oksanen2022anoppi, Kleinberg2022textwash}. The advent of LLMs has initiated a paradigm shift in this domain. LLMs are now employed as potent, context-aware anonymization tools capable of zero-shot PII removal and replacement \citep{Staab2024largeICLR, Liu2023deidgpt_arxiv}. This development is complemented by methodologies that distill the knowledge from large, proprietary models into smaller, open-source versions for practical deployment \citep{deusser-etal-2025-resource}. Conversely, the inferential capabilities of LLMs also position them as formidable adversaries in re-identifying individuals from inadequately anonymized text, thereby elevating the requirements for robust privacy measures \citep{Patsakis2023man}. This dual capacity of LLMs demands a more profound comprehension of their interplay with anonymized data.

Empirical evaluations of anonymization effects have mostly been domain-specific. In medicine, \citet{Larbi2023clinical} found variable performance losses across five clinical NLP tasks and persistent re-identification risk. In contrast, \citet{Vakili2024pseudonymization} reported that pseudonymization before training Swedish Clinical BERT resulted in near-identical downstream performance. Financial-domain work combining Differential Privacy with Federated Learning explicitly highlights the privacy-utility trade-offs in transformer-based text classification \citep{basu-etal-2021-privacy}. Meanwhile, analyses of pseudonymization strategies reveal that seemingly minor design differences can have a significant impact on classification and summarization outcomes \citep{Yermilov2023privacy}.

The rapid evolution of privacy-preserving techniques has created a pressing need for robust evaluation frameworks. \citet{ai6020029} introduced Priv-IQ, a comprehensive multimodal benchmark measuring LLM privacy intelligence across eight competencies including visual privacy, multilingual capabilities, and knowledge of privacy law. \citet{sun2025effectiveness} presented a systematic evaluation framework for privacy-preserving algorithms across different LLM architectures (Mistral-7B, Llama2-7b, Falcon-7b), examining three scenarios: protecting training data only, user queries only, or both. 

Domain-specific evaluation tools have also emerged. \citet{zhang2024safetybench} introduced SafetyBench with over 11,000 questions across seven safety categories including privacy concerns, while \citet{andriushchenko2025agentharm} presented AgentHarm for evaluating harmful LLM agent behaviors including privacy violations. These specialized benchmarks complement general privacy evaluations by capturing nuanced risks in specific deployment contexts.

For a more complete overview on the topic of textual anonymization, we refer the interested reader to our survey paper on various anonymization techniques and recent advances \citep{deusser2025asurvey}.

While these studies offer vital insights, they are often confined to specific domains (e.g., clinical or financial texts), employ a restricted set of NLP tasks (most commonly classification), or are predicated on earlier model architectures such as BERT. Our research broadens this line of inquiry by conducting a large-scale, systematic empirical evaluation with a more extensive scope. We evaluate a diverse set of five modern and highly capable LLMs, with parameter counts ranging from 7 to 72 billion, across eleven distinct benchmarks that encompass a wide array of reasoning and linguistic capabilities. This allows us to transcend domain-specific findings and identify more generalizable patterns concerning the privacy-utility trade-off. We aim to establish a cross-domain baseline for understanding how input anonymization affects the performance of contemporary LLMs.

\section{Methodology}

This section details the methodology employed to investigate the impact of input anonymization on the performance of LLMs. We describe our experimental pipeline, the selection criteria for benchmarks and models, and the specific experimental setup used to compare performance on original versus anonymized inputs.

\subsection{Pipeline}
To systematically evaluate the effect of anonymization, we developed a modular pipeline for our experiments. This pipeline facilitates the processing of diverse benchmarks with various LLMs, both with and without anonymization. The core stages are:

\begin{enumerate}
    \item \textbf{Data preparation}: Benchmarks are downloaded, preprocessed, and sampled as required. Each benchmark is encapsulated, storing prompt templates, questions, and reference answers.
    \item \textbf{Anonymization}: Questions (and associated prompt components, if applicable) are processed by an anonymizer, which replaces personally identifiable information (PII) with placeholders. A mapping between original and anonymized entities is retained.
    \item \textbf{Model inference}: Selected LLMs generate responses to both the original and anonymized versions of the benchmark questions.
    \item \textbf{De-anonymization (for evaluation)}: For anonymized outputs, placeholders in the LLM-generated responses are mapped back to their original PII using the stored mappings. This step is crucial for ensuring fair comparison with reference answers.
    \item \textbf{Evaluation}: LLM responses (with original PII restored for anonymized versions) are assessed against reference answers using the predefined evaluation metrics specific to each benchmark.
    \item \textbf{Results aggregation}: Performance scores are collated, and metrics such as the relative performance change between anonymized and non-anonymized runs are computed.
\end{enumerate}

\begin{figure}
  \centering
  \includegraphics[width=\textwidth]{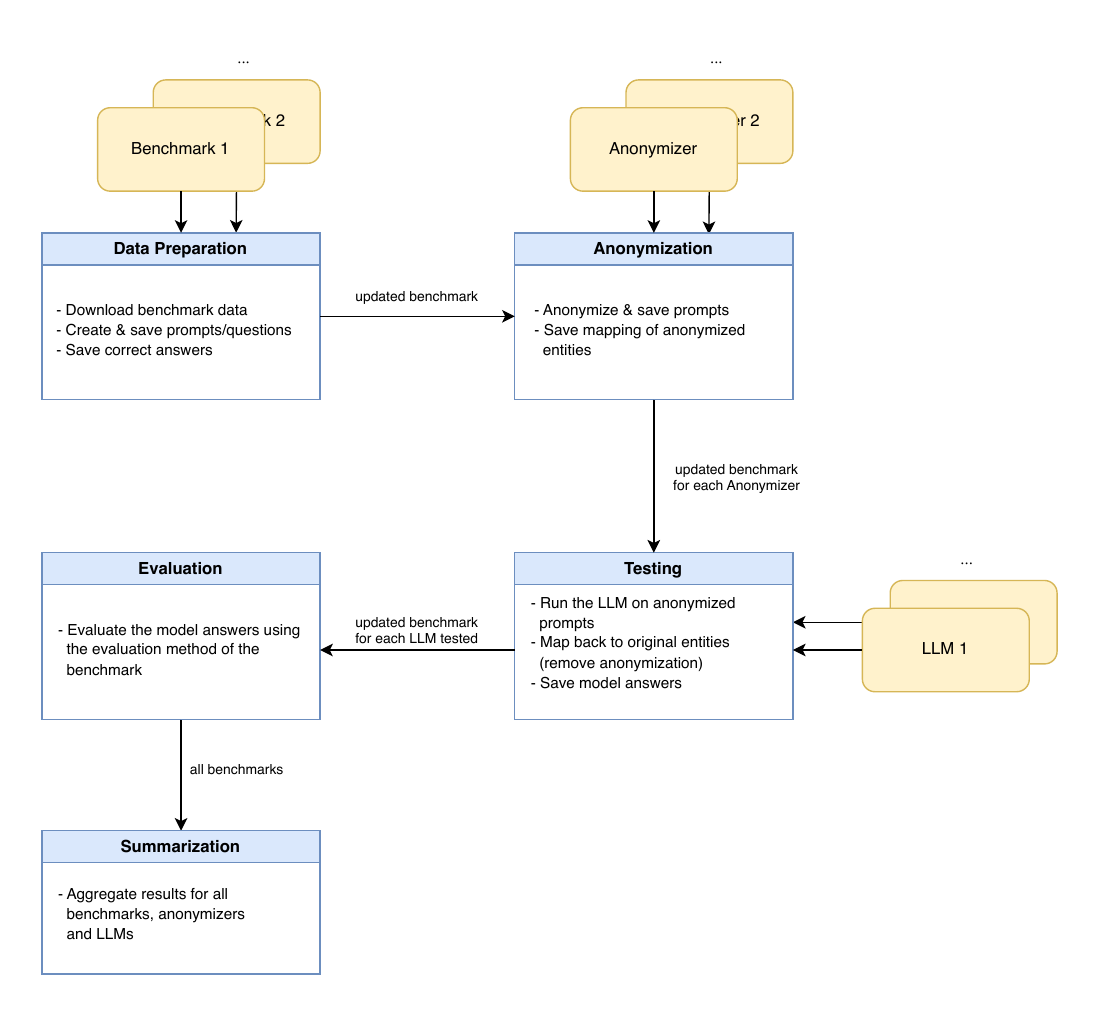} 
  \caption{Architecture overview of our encoder-only contradiction detection approach.}
  \label{fig:anon_benchmark_pipeline}
\end{figure}

Figure \ref{fig:anon_benchmark_pipeline} provides a conceptual overview of this pipeline. The design emphasises extensibility, allowing for the straightforward integration of new benchmarks, anonymization strategies, and LLMs through abstract base classes. 

\subsection{Dataset and Benchmark Selection}
To evaluate the impact of anonymization across diverse linguistic and cognitive tasks, we selected eleven established benchmarks. Our selection was guided by several criteria: 
\begin{enumerate}
    \item \textbf{Popularity and recognition}: Widespread use within the research community.
    \item \textbf{Openness}: Publicly available test sets and evaluation methodologies to ensure transparency and reproducibility.
    \item \textbf{Language coverage}: Predominantly English-language benchmarks, with the inclusion of one German-English translation task (WMT 2014) to assess cross-lingual effects.
    \item \textbf{Task Diversity}: A broad range of domains (e.g., science, medicine, commonsense reasoning) and prompt styles (e.g., zero-shot, few-shot, chain-of-thought) to avoid a narrow focus.
\end{enumerate}
Domains such as coding and mathematics were excluded, as they typically do not involve PII, meaning anonymization would have minimal impact. We identified candidate benchmarks by consulting resources such as the Open LLM Leaderboard \citep{open-llm-leaderboard-v2} and the Stanford HELM project \citep{liang2023holistic}. The selected benchmarks are detailed in Table~\ref{tab:anon_benchmark_overview}.

\begin{table}
    \centering 
    \footnotesize
    \caption{Overview of selected benchmarks: Domains, prompt type, and references. CoT denotes Chain-of-Thought; RAG denotes Retrieval-Augmented Generation.}
    \label{tab:anon_benchmark_overview}
    \begin{tabular}{lp{4.6cm}ll} 
    \toprule
    Benchmark & Domain & Prompt Type  & Reference\\
    \midrule
    ARC & Grade-school science reasoning & Zero-Shot & \cite{clark2018thinksolvedquestionanswering} \\
    \rowcolor{gray!10}BIG-Bench Hard & Challenging reasoning tasks & CoT, Few-Shot & \cite{suzgun2022challengingbigbenchtaskschainofthought}\\
    EQ-Bench & Emotional intelligence & Zero-Shot, Self-Critique & \cite{paech2024eqbenchemotionalintelligencebenchmark}\\
    \rowcolor{gray!10}HellaSwag & Commonsense reasoning & Zero-Shot & \cite{zellers2019hellaswagmachinereallyfinish} \\
    IFEval & Complex instruction following & Zero-Shot & \cite{zhou2023instructionfollowingevaluationlargelanguage} \\
    \rowcolor{gray!10}MedQA & Medical knowledge \& reasoning & Zero-Shot & \cite{jin2021disease}\\
    MMLU-PRO & Multitask academic \& professional knowledge & CoT, Few-Shot & \cite{wang2024mmluprorobustchallengingmultitask}\\
    \rowcolor{gray!10}MUSR & Multi-step reasoning & CoT, Zero-Shot & \cite{sprague2024musrtestinglimitschainofthought} \\
    RGB & Retrieval-Augmented Generation & Zero-Shot & \cite{chen2024benchmarking}\\
    \rowcolor{gray!10}TruthfulQA & Truthfulness \& debunking misconceptions & Zero-Shot & \cite{lin-etal-2022-truthfulqa} \\ 
    WMT 2014 (En-De) & Machine translation (English to German)  & Few-Shot & \cite{bojar2014findings}\\ 
    \bottomrule
\end{tabular}    
\end{table}

Given computational constraints associated with evaluating multiple LLMs across numerous benchmarks, we standardised the number of samples. For most benchmarks, a maximum of 500 samples were randomly selected from their respective test sets. For benchmarks comprising multiple tasks (specifically, BIG-Bench Hard and MMLU-PRO), this cap was increased to 1,000 samples to ensure adequate representation per subtask. This resulted in a total of 6,210 unique samples being evaluated. An overview of original and adjusted sample sizes can be found in Table~\ref{tab:anon_dataset_size_summary_table}.

\begin{table}
    \centering
    \footnotesize
    \caption{Benchmark summary: Number of tasks, original sample counts, adjusted sample counts for this study, and average input length (average number of tokens) of adjusted samples.}
    \label{tab:anon_dataset_size_summary_table}
    \begin{tabular}{lrrrr}
\toprule
Benchmark     & Subtasks & Sample Counts & Adj. Sample Counts & Avg. Length\\ 
\midrule
ARC                  & 2                          & 3548           & 500    & 95.79 \\ 
\rowcolor{gray!10}BIG-Bench Hard         & 27                         & 6421           & 999    & 812.11 \\ 
EQ-Bench              & 2                          & 342             & 342   & 404.76 \\ 
\rowcolor{gray!10}HellaSwag            & 1                          & 10042          & 500    & 133.98 \\ 
IFEval               & 1                          & 541             & 500   & 44.02  \\ 
\rowcolor{gray!10}MedQA                & 1                          & 1273           & 500    & 170.37 \\  
MMLU-PRO              & 13                         & 11032          & 994   & 382.70 \\ 
\rowcolor{gray!10}MUSR                 & 3                          & 375              & 375  & 1295.43   \\
RGB                  & 3                          & 500             & 500   & 921.27  \\
\rowcolor{gray!10}TruthfulQA           & 1                          & 790             & 500   & 144.87  \\ 
WMT 2014 (En-De)      & 1                          & 3003           & 500    & 44.74 \\ 
\midrule
Total               &  55                      &  38967               & 6210 & 417.62  \\ 
\bottomrule
\end{tabular}
\end{table}

When official prompts were available with the benchmarks, they were utilised. If multiple official prompts existed (e.g., zero-shot vs. few-shot), we selected the one anticipated to yield stronger baseline performance, typically favouring few-shot prompts. For benchmarks providing only datasets without predefined prompts, we first sought established prompt implementations (e.g., from the HELM repository). If no suitable reference was found, we created minimal zero-shot prompts. Importantly, for consistency, any prompt components, including few-shot examples, were anonymized alongside the primary question input to ensure comprehensive PII removal from the text presented to the LLM. This approach is crucial as our focus is on the relative impact of anonymization rather than achieving absolute state-of-the-art scores.

\subsection{Model Selection}
Our selection of LLMs, as seen in Table \ref{tab:anon_llm_overview}, aims for diversity in terms of model providers, parameter sizes, and architectures, whilst remaining within feasible computational limits for extensive experimentation. The primary objective is to analyse the \emph{relative} change in performance due to anonymization, rather than to benchmark absolute model capabilities against the state-of-the-art.

\begin{table}
    \centering
    \footnotesize
    \caption{Overview of large language models (LLMs) evaluated in this study. Shortened names are used throughout the paper for brevity.}
    \label{tab:anon_llm_overview}
    \begin{tabular}{llllc}
    \toprule
        Short Name & Full Model Name & Developer & Reference & Parameter Count  \\ 
        \midrule
         GPT-4o mini & GPT-4o mini & OpenAI & \citet{OpenAIGPT4omini2024} & UNKNOWN \\ 
         \rowcolor{gray!10}Teuken-7B & Teuken-7B-instruct-research-v0.4 & OpenGPT-X & \citet{ali2024teuken} & 7B \\ 
         Llama-3.1-8B & Llama-3.1-8B-Instruct & Meta & \citet{meta_llama_3_2024} & 8B \\
         \rowcolor{gray!10}Gemma-2-27B & Gemma-2-27b & Google & \citet{gemmateam2024gemma2improvingopen} & 27B \\
         Qwen2.5-72B & Qwen2.5-72B-Instruct & Alibaba & \citet{qwen2025qwen25technicalreport} & 72B \\
    \bottomrule
\end{tabular}
\end{table}

\subsection{Experimental Setup}
The core experiment compares the performance of each selected LLM on each benchmark under two conditions: (i) original, non-anonymized inputs, and (ii) anonymized inputs.

\subsubsection{Anonymization Configuration}
\label{subsubsec:anonymization_configuration}

For the primary experiments, we employ a pseudonymization strategy implemented by our anonymization tool \citep{deusser-etal-2025-resource}. This tool is a sophisticated LLM-based system that performs named entity recognition (NER) to identify PII and subsequently replaces them. Specifically, entities are substituted with a category label and a unique, consistent identifier; for example, every occurrence of "London" within a document is replaced by "<LOC>-1", while "Berlin" would become "<LOC>-2". This ensures that the uniqueness of entities is preserved, which is crucial for downstream tasks that rely on entity distinctions. The underlying model is a fine-tuned version of a smaller, efficient language model, optimized for a high recall score. In our evaluations, this tool achieves an F1-score of over 88\% and a recall of over 91\%.

\subsubsection{Comparison of Anonymization Techniques}
\label{subsubsec:anonymization_comparison_anonymization_techniques}
To investigate the impact of different anonymization approaches, we evaluate a subset of models and benchmarks using four distinct techniques in addition to pseudonymization:
\begin{itemize}
    \item \textbf{Generalization}: Replacing specific PII with broader category labels (e.g., \say{London} becomes \say{[CITY]}).
    \item \textbf{Masking}: Parts of entities are obscured in a consistent pattern, preserving some surface information (e.g., \say{London} becomes \say{Lo****}).
    \item \textbf{Randomization}: Entities are substituted with fixed but randomly generated strings that remain consistent across the dataset (e.g., \say{London} becomes \say{Xj92Lp})
    \item \textbf{Redaction}: Replacing PII with a fixed, generic marker (e.g., \say{[REDACTED]} or \rule{1ex}{1ex}\rule{1ex}{1ex}\rule{1ex}{1ex}\rule{1ex}{1ex}\rule{1ex}{1ex}).
\end{itemize}
This comparative analysis is performed on GPT-4o mini using the MedQA, MUSR, and RGB benchmarks.

\subsubsection{Impact of Explicit Anonymization Prompts}
\label{subsubsec:anonymization_explicit_anonymization_prompts}
We also explore whether explicitly informing the LLM that an input has been anonymized affects performance. For this, two prefixes are prepended to anonymized prompts for GPT-4o mini on the MedQA, MUSR, and RGB benchmarks:
\begin{itemize}
    \item Short Prefix: \say{This prompt is anonymized. Some context may be missing; respond as accurately as possible.}
    \item Long Prefix: \say{This prompt has been anonymized by removing all sensitive entities. Some contextual details may be missing, but please generate the most accurate and effective response possible based on the available information.}
\end{itemize}
Performance is compared against anonymized inputs without any prefix and non-anonymized inputs.

\section{Experiments}

This section presents the empirical evaluation of LLM performance under anonymization. We first detail the main experimental setup and results concerning our primary pseudonymization strategy. Subsequently, we investigate the effects of varying anonymization techniques and the utility of explicitly informing models about input anonymization.

All experiments are conducted on a system equipped with four NVIDIA Tesla V100 GPUs, each with 32GB of memory. To ensure reproducibility of results, a global random seed is used for data sampling (where applicable for benchmarks) and any stochastic processes within the models if not entirely avoidable. Crucially, the sampling temperature for LLM generation is set to 0 for all models and tasks, aiming for as deterministic as possible outputs.

\subsection{Impact of Pseudonymization on LLM Performance}
\label{ssec:main_experiment_results}

We evaluated the five LLMs detailed in Table~\ref{tab:anon_llm_overview} (GPT-4o mini, Teuken-7B, Llama-3.1-8B-Instruct, Gemma-2-27B, and Qwen2.5-72B) across the eleven benchmarks described in Table~\ref{tab:anon_benchmark_overview}. For each benchmark, models processed both original and anonymized inputs. The anonymization employed the pseudonymization strategy outlined in Section~\ref{subsubsec:anonymization_configuration} (from \citet{deusser-etal-2025-resource}), replacing PII with category-unique placeholders (e.g., \say{London} to \say{<LOC>-1}). Performance was measured using the standard metrics for each benchmark. 

\begin{table}
\centering
\footnotesize
\caption{Overview of the benchmarks: Domain and Promptstyle}
\begin{tabular}{lp{4.6cm}ll} 
    \toprule
    Benchmark & Domain & Prompt Type  & Reference\\
    \midrule
    ARC & Grade-school science reasoning & Zero-Shot & \cite{clark2018thinksolvedquestionanswering} \\
    \rowcolor{gray!10}BIG-Bench Hard & Challenging reasoning tasks & CoT, Few-Shot & \cite{suzgun2022challengingbigbenchtaskschainofthought}\\
    EQ-Bench & Emotional intelligence & Zero-Shot, Self-Critique & \cite{paech2024eqbenchemotionalintelligencebenchmark}\\
    \rowcolor{gray!10}HellaSwag & Commonsense reasoning & Zero-Shot & \cite{zellers2019hellaswagmachinereallyfinish} \\
    IFEval & Complex instruction following & Zero-Shot & \cite{zhou2023instructionfollowingevaluationlargelanguage} \\
    \rowcolor{gray!10}MedQA & Medical knowledge \& reasoning & Zero-Shot & \cite{jin2021disease}\\
    MMLU-PRO & Multitask academic \& professional knowledge & CoT, Few-Shot & \cite{wang2024mmluprorobustchallengingmultitask}\\
    \rowcolor{gray!10}MUSR & Multi-step reasoning & CoT, Zero-Shot & \cite{sprague2024musrtestinglimitschainofthought} \\
    RGB & Retrieval-Augmented Generation & Zero-Shot & \cite{chen2024benchmarking}\\
    \rowcolor{gray!10}TruthfulQA & Truthfulness \& debunking misconceptions & Zero-Shot & \cite{lin-etal-2022-truthfulqa} \\ 
    WMT 2014 (En-De) & Machine translation (English to German)  & Few-Shot & \cite{bojar2014findings}\\ 
    \bottomrule
\end{tabular}
\label{tab:anon_benchmarks_overview}
\end{table}


Table~\ref{tab:anon_main_results_overview} presents the performance scores for all models on all benchmarks, under both original and anonymized conditions. Several patterns emerge from this comprehensive evaluation. First, we observe that model performance is not uniformly affected across benchmarks. While some tasks show minimal degradation (e.g., ARC, HellaSwag, IFEval with drops typically under 5 percentage points), others experience substantial losses (e.g., RGB with drops of 30-60 percentage points for most models). 

Second, the data reveals interesting model-specific behaviors. Notably, Teuken-7B achieves a score of 0.00 on both ARC and HellaSwag regardless of anonymization. This suggests fundamental limitations in this model's ability to handle these specific task formats, independent of entity information. In contrast, Qwen2.5-72B and GPT-4o mini demonstrate strong baseline performance across most benchmarks, achieving scores above 0.60 on 8 out of 11 tasks in the non-anonymized condition. However, this superior baseline performance comes with a trade-off: these same models experience the largest absolute performance drops when anonymization is applied.

Third, TruthfulQA stands out as the only benchmark where all five models either maintain or improve performance under anonymization. This counterintuitive result warrants closer examination, which we provide in our subsequent analysis. The consistency of this improvement across models of varying architectures and capabilities suggests a systematic phenomenon rather than a random fluctuation.

\begin{figure}
  \centering
  \includegraphics[width=0.95\textwidth]{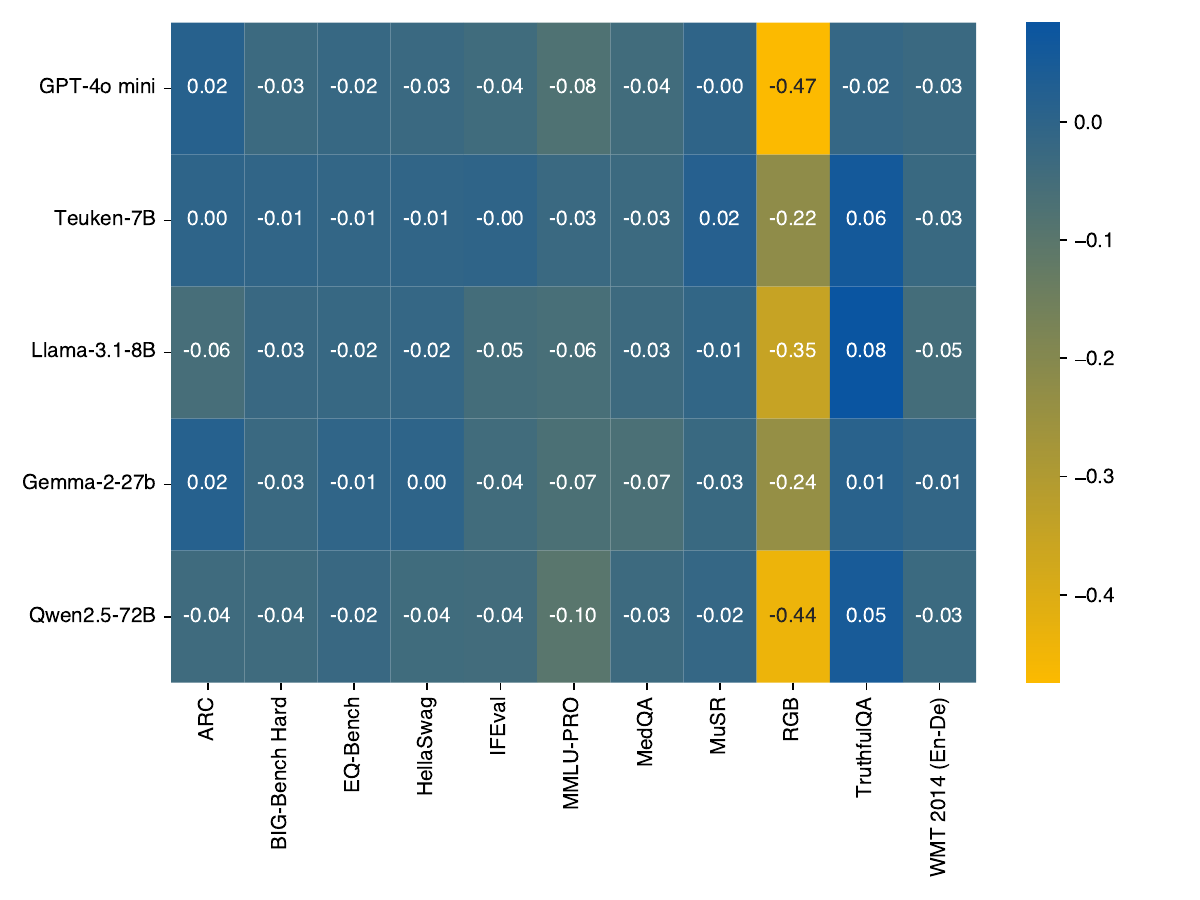} 
  \vspace{17pt}
  \caption{Difference in benchmark performance scores (\emph{anonymized} minus \emph{non-anonymized}) across multiple language models and tasks. Negative values indicate performance drops due to anonymization.}
  \label{fig:anon_performance_difference_heatmap}
\end{figure}

Figure~\ref{fig:anon_performance_difference_heatmap} illustrates the performance difference (anonymized minus non-anonymized scores) as a heatmap, providing a visual representation of anonymization impact across the model-benchmark matrix. The color intensity corresponds to the magnitude of performance change, with darker blue indicating smaller drops (or improvements) and yellow/orange indicating larger performance losses.

Several insights emerge from this visualization. First, RGB exhibits the most severe and consistent performance degradation across all models, with drops ranging from -0.22 (Teuken-7B) to -0.47 (GPT-4o mini). This catastrophic decline suggests that RGB's retrieval-augmented generation tasks are fundamentally dependent on specific entity information that pseudonymization clearly obscures. The task requires models to integrate information from multiple sources while filtering out noise, and replacing concrete entities with abstract placeholders appears to disrupt this integration process severely.

Interestingly, the positive values for TruthfulQA represent a striking anomaly. This consistent improvement across multiple models suggests that anonymization may actually help models avoid factual errors in this specific context. We hypothesize that TruthfulQA questions often contain named entities that trigger incorrect memorized associations. By replacing \say{Which country in Europe...} with \say{Which country in <LOC>-1...}, we may prevent models from confidently generating plausible-sounding but incorrect facts tied to specific names they encountered during training.

\begin{table}
\centering
\footnotesize
\caption{Performance overview of the 5 tested LLMs across all benchmarks (anonymized and non-anonymized). Higher scores indicate better performance. The best scores are bold.}
\begin{tabular}{l c c c c c c c}
    \toprule
    Benchmark & Anonymized & GPT-4o mini & Teuken-7B & Llama-3.1-8B & Gemma-2-27B & Qwen2.5-72B \\
    \midrule
    
    ARC & no & 0.62 & 0.00 & 0.84 & 0.29 & \textbf{0.95} \\
    & yes & 0.64 & 0.00 & 0.79 & 0.30 & \textbf{0.91} \\
    
    \rowcolor{gray!10}BIG-Bench Hard & no & 0.24 & 0.18 & 0.23 & \textbf{0.25} & 0.23 \\
    \rowcolor{gray!10}& yes & 0.21 & 0.18 & 0.20 & \textbf{0.22} & 0.19 \\
    
    EQ-Bench & no & 0.74 & 0.30 & 0.66 & \textbf{0.76} & 0.72 \\
    & yes & 0.72 & 0.29 & 0.64 & \textbf{0.76} & 0.70 \\
     
    \rowcolor{gray!10}HellaSwag & no & 0.77 & 0.01 & 0.50 & 0.00 & \textbf{0.82} \\
    \rowcolor{gray!10}& yes & 0.74 & 0.00 & 0.49 & 0.00 & \textbf{0.78} \\
     
    IFEval & no & 0.81 & 0.28 & 0.75 & 0.62 & \textbf{0.83} \\
    & yes & 0.77 & 0.27 & 0.70 & 0.57 & \textbf{0.79} \\

    \rowcolor{gray!10}MedQA & no & 0.88 & 0.37 & 0.91 & \textbf{0.94} & 0.87 \\
    \rowcolor{gray!10}& yes & 0.83 & 0.34 & 0.88 & \textbf{0.87} & 0.83 \\
     
    MMLU-PRO & no & 0.63 & 0.21 & 0.47 & 0.55 & \textbf{0.71} \\
    & yes & 0.55 & 0.18 & 0.42 & 0.49 & \textbf{0.61} \\
    
    \rowcolor{gray!10}MUSR & no & 0.70 & 0.52 & 0.63 & 0.62 & \textbf{0.75} \\
    \rowcolor{gray!10}& yes & 0.70 & 0.54 & 0.62 & 0.59 & \textbf{0.73} \\
    
    RGB & no & 0.80 & 0.53 & 0.69 & 0.63 & \textbf{0.82} \\
    & yes & 0.32 & 0.31 & 0.34 & \textbf{0.39} & 0.38 \\

    \rowcolor{gray!10}TruthfulQA & no & 0.61 & 0.45 & \textbf{0.70} & 0.67 & 0.69 \\
    \rowcolor{gray!10}& yes & 0.60 & 0.51 & \textbf{0.78} & 0.68 & 0.74 \\
    
    WMT 2014 (En-De) & no & \textbf{0.30} & 0.26 & 0.24 & 0.17 & \textbf{0.30} \\
    & yes & \textbf{0.27} & 0.24 & 0.19 & 0.16 & \textbf{0.27} \\
     
    \bottomrule
\end{tabular}
\label{tab:anon_main_results_overview}
\end{table}

\begin{figure}
  \centering
  \includegraphics[width=0.9\textwidth]{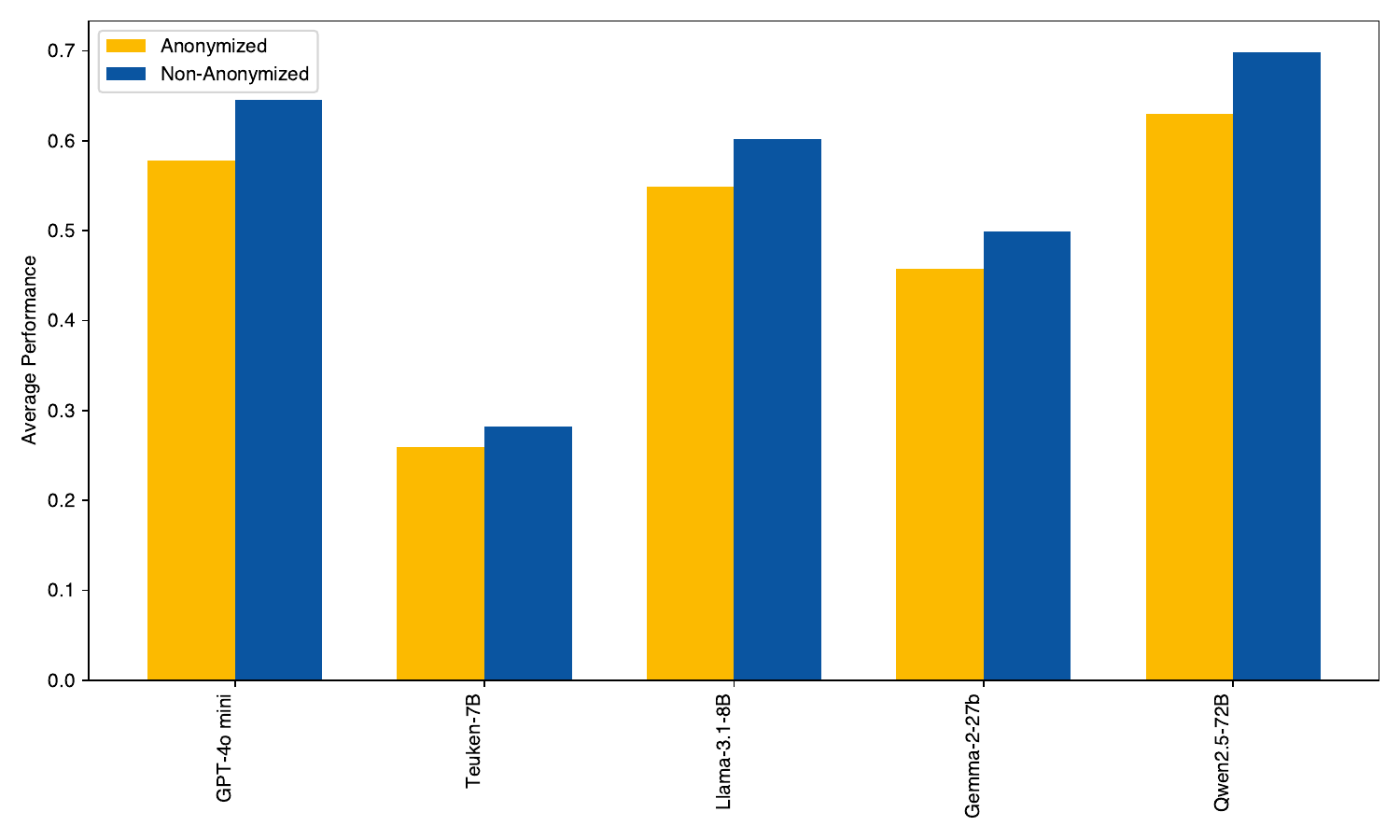} 
  \vspace{17pt}
  \caption{Average performance of the five different models under anonymized and non-anonymized conditions. For all models tested, the non-anonymized version shows equal or slightly higher average performance.}
  \label{fig:anon_avg_perf_model}
\end{figure}

Figure~\ref{fig:anon_avg_perf_model} displays the average performance across all benchmarks for each model. While Qwen2.5-72B and GPT-4o mini achieve the highest average scores, they also exhibit the largest average performance degradation when inputs are anonymized. This suggests that more capable models might rely more heavily on specific entity information, making them more susceptible to the information loss caused by anonymization. For instance, Qwen2.5-72B's average performance drops by 6.9 percentage points, while Teuken-7B, the lowest-performing model, sees a smaller average drop of 2.3 percentage points.

The impact of anonymization varies considerably across benchmarks. Notably, TruthfulQA is the only benchmark where performance consistently \emph{improves} with anonymization for most models (e.g., an increase of  8 percentage points for Llama-3.1). This suggests that removing named entities might reduce the model's tendency to hallucinate or retrieve incorrect factual associations for this specific task.

Conversely, the RGB benchmark, which tests retrieval-augmented generation capabilities like noise robustness and information integration, suffers the most significant performance degradation. For example, GPT-4o mini's score on RGB drops by approximately 48 percentage points (from 0.80 to 0.32). This substantial decrease highlights RGB's reliance on precise entity information for its complex reasoning and information synthesis requirements.

\begin{table}[h!]
\centering
\footnotesize
\caption{Average entity statistics across all benchmarks. For each benchmark, we report: (1) the average number of total entities (including repetitions), (2) the average number of unique entities, and (3) the average length of each input (in tokens).}
\begin{tabular}{lccc} 
    \toprule
    Benchmark & Avg. Total Entities & Avg. Unique Entities & Avg. Length \\
    \midrule
    ARC & 0.83 & 0.61 & 97.65 \\
    \rowcolor{gray!10}BIG-Bench Hard & 52.43 & 10.15 & 840.92 \\
    EQ-Bench & 17.30 & 2.75 & 405.27 \\
    \rowcolor{gray!10}HellaSwag & 0.95 & 0.61 & 135.32 \\
    IFEval & 0.67 & 0.61 & 44.94 \\
    \rowcolor{gray!10}MedQA & 2.53 & 2.27 & 174.54 \\
    MMLU-PRO & 7.43 & 4.80 & 395.25 \\
    \rowcolor{gray!10}MUSR & 67.47 & 8.69 & 1317.20 \\
    RGB & 79.49 & 35.26 & 1000.23 \\
    \rowcolor{gray!10}TruthfulQA & 12.68 & 7.67 & 170.99 \\
    WMT 2014 & 4.20 & 2.10 & 55.83 \\
    \bottomrule
\end{tabular}
\label{tab:entity_token_stats}
\end{table}

Table~\ref{tab:entity_token_stats} provides statistics on the average number of entities (total and unique) and input lengths across benchmarks, which help contextualize the varying impact of anonymization. While benchmarks like RGB and MUSR exhibit high entity counts and long inputs, only RGB experiences a substantial performance drop after anonymization. MUSR, despite similarly high entity statistics, shows minimal performance degradation. Conversely, benchmarks with few entities and short inputs, such as HellaSwag and IFEval, display only small performance drops, aligning with expectations. However, benchmarks like MMLU-PRO and MedQA show noticeable degradation despite moderate entity counts, suggesting that entity quantity alone does not determine impact. To investigate this further, we conducted correlation analyses between performance drops and each of the reported metrics (total entities, unique entities, and input length), but found no statistically significant correlations. This indicates that anonymization effects are likely mediated by task-specific dependencies on entity semantics and context, rather than raw input characteristics.

\subsection{Comparing Different Anonymization Techniques}

We compare five anonymization techniques (redaction, generalization, pseudonymization, randomization, and masking), as described in Section~\ref{subsubsec:anonymization_comparison_anonymization_techniques}. This experiment is conducted using GPT-4o mini on a subset of three benchmarks: MedQA, MUSR, and RGB, chosen for their diverse characteristics.

\begin{figure}
  \centering
  \includegraphics[width=0.95\textwidth]{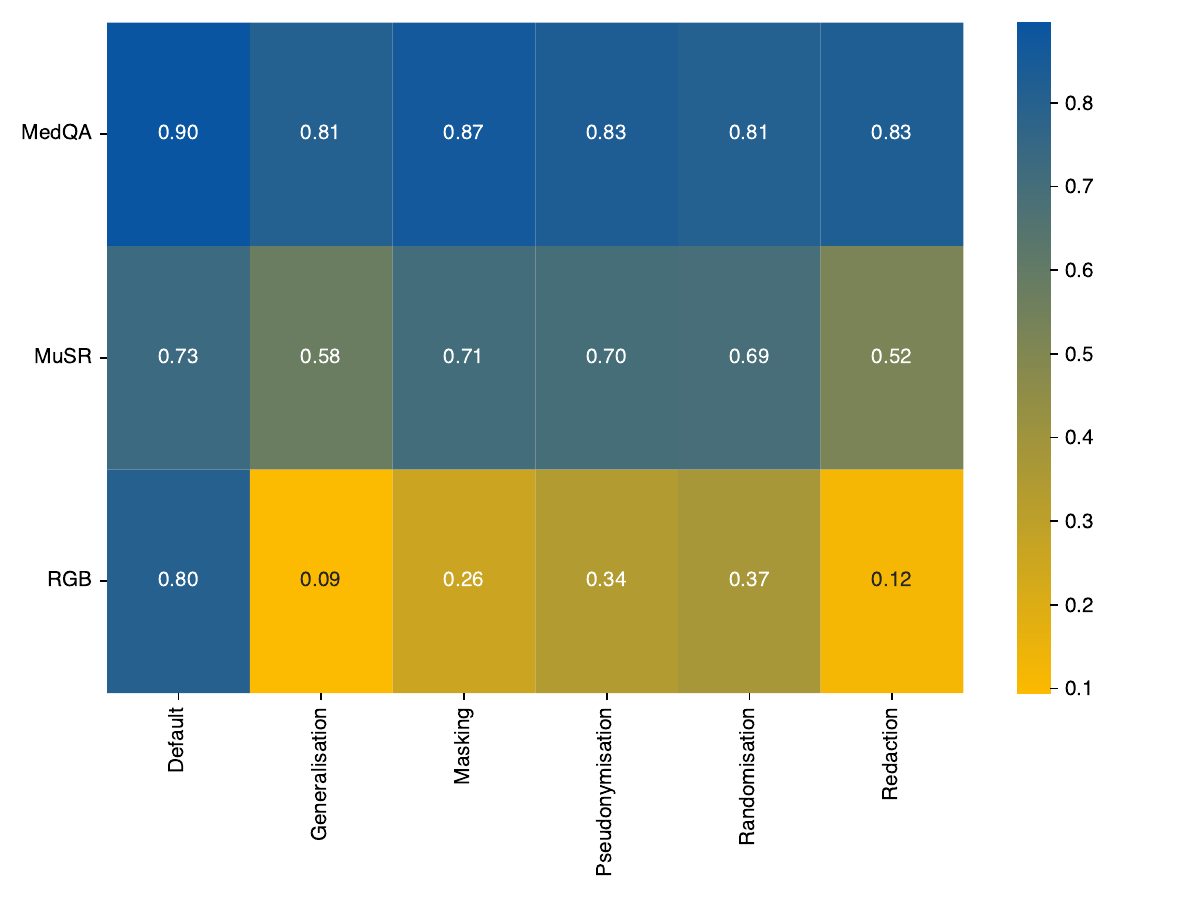} 
  \vspace{17pt}
  \caption{Performance of GPT-4o mini on MedQA, MUSR, and RGB, comparing redaction, generalization, pseudonymization, randomization, and masking.}
  \label{fig:anon_techniques_comparison}
\end{figure}

Figure~\ref{fig:anon_techniques_comparison} shows the performance of GPT-4o mini under different anonymization methods. Reversible techniques (pseudonymization, randomization, and masking), which allow for a potential one-to-one mapping back to original entities, generally preserve performance better than irreversible ones (redaction, generalization).

Generalization (e.g., \say{London} $\rightarrow$ \say{[CITY]}) and Redaction (e.g., \say{London} $\rightarrow$ \say{\rule{1ex}{1ex}\rule{1ex}{1ex}\rule{1ex}{1ex}\rule{1ex}{1ex}\rule{1ex}{1ex}}) consistently resulted in the largest performance drops. For instance, on MuSR, Generalization led to a score of 0.58 and Redaction 0.52, compared to 0.70 for pseudonymization and 0.73 for the original input. 
This is likely because these methods remove too much information, making it difficult for the model to establish necessary entity distinctions and for evaluation scripts, which expect original entities after de-anonymization, to score correctly.

Masking (e.g., \say{London} $\rightarrow$ \say{Lo****}) and Randomization (e.g., \say{London} $\rightarrow$ \say{Xj92Lp}) performed comparably to or slightly worse than Pseudonymization. Masking performed well on MedQA and MUSR but saw a significant drop on RGB, possibly due to the nature of information required by RGB that simple partial string matching cannot satisfy. 

These results underscore that techniques preserving unique identifiers for distinct entities, even if abstract, are preferable for maintaining utility.

\subsection{Impact of Explicit Anonymization Prompts}
\label{anoynmization_prompts}

Here, we investigate whether explicitly informing the LLM about input anonymization has an impact on performance. Using GPT-4o mini once again on MedQA, MUSR, and RGB, we prepend two types of instructional prefixes (short and long, as detailed in Section~\ref{subsubsec:anonymization_explicit_anonymization_prompts}) to anonymized inputs and compared results against anonymized inputs without a prefix and original inputs.

\begin{figure}
  \centering
  \includegraphics[width=0.95\textwidth]{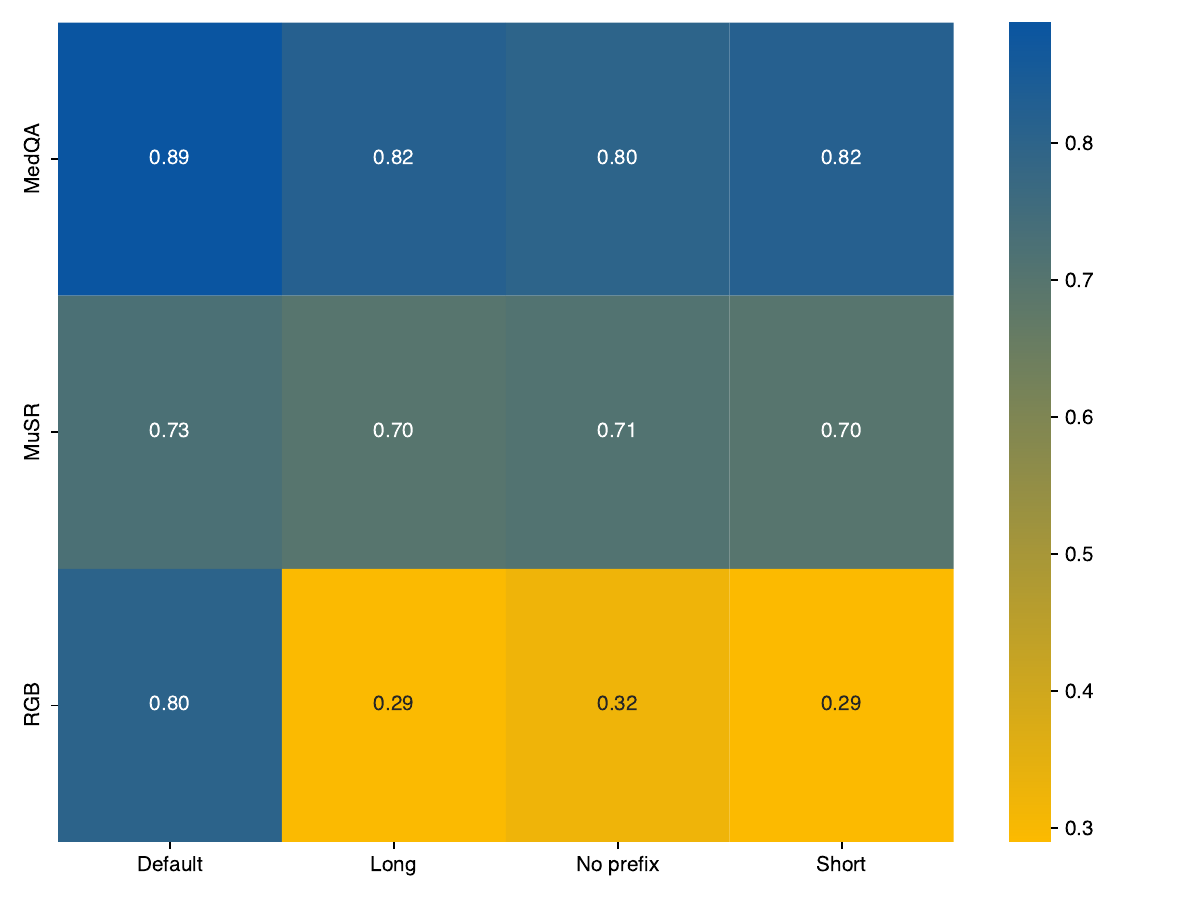} 
  \vspace{17pt}
  \caption{Performance of GPT-4o mini on MedQA, MUSR, and RGB with original inputs, anonymized inputs (no prefix), and anonymized inputs with short or long explicit prefixes.}
  \label{fig:anon_prefixes_comparison}
\end{figure}

Figure~\ref{fig:anon_prefixes_comparison} summarises the findings. Adding an explicit textual cue that the input has been anonymized did not yield any significant or consistent improvement in LLM performance across the tested benchmarks and prefixes. The performance scores were very similar regardless of whether a short prefix, a long prefix, or no prefix was used with the anonymized input. This suggests that current LLMs do not substantially alter their reasoning strategy based on such meta-information, and the performance impact is primarily driven by the modification of content itself.

\subsection{Benchmark Examples}

In this subsection, we explore why we observe the performance differences shown in Table~\ref{tab:anon_main_results_overview}, Figure~\ref{fig:anon_performance_difference_heatmap} to \ref{fig:anon_techniques_comparison}, and described in the previous subsections by investigating various examples from the test sets of our studied datasets. Note that some questions and answers may have been shortened for readability in Tables~\ref{tab:example1} to \ref{tab:example5}.

\begin{table}[t]
    \centering
    \caption{Example 1 (from the ARC Benchmark)}
    \footnotesize
    \begin{tabularx}{\textwidth}{
    >{\raggedright\arraybackslash}X
    >{\raggedright\arraybackslash}X
}
\toprule
\rowcolor{gray!20} 
\textbf{Original Case} & \textbf{Anonymized Case} \\
\midrule

\textbf{Question:} A solar eclipse occurs when \newline
A. the Moon blocks the Earth from the Sun. \newline
B. the first four planets in the solar system are aligned. \newline
C. Earth blocks the Moon from the Sun. \newline
D. Earth's shadow falls on the Sun.
\vspace{4mm} \newline
\textbf{Predicted Answer:} A \hfill \textbf{Score:} 1.0

& 

\textbf{Question:} A solar eclipse occurs when \newline
A. the Moon blocks the <LOC>-1 from the <LOC>-2. \newline
B. the first four planets in the solar system are aligned. \newline
C. <LOC>-1 blocks the Moon from the <LOC>-2. \newline
D. <LOC>-1's shadow falls on the <LOC>-2.
\vspace{4mm} \newline
\textbf{Predicted Answer:} D \hfill \textbf{Score:} 0.0
\\
\midrule 

\rowcolor{gray!10} 
\multicolumn{2}{p{\dimexpr\textwidth-2\tabcolsep}}{\textbf{Reference Answer:} A} \\

\bottomrule
\end{tabularx}
    \label{tab:example1}
\end{table}

In the question about solar eclipses depicted in Table~\ref{tab:example1}, our anonymization tool replaced ``Earth'' and ``Sun'' with placeholders like \texttt{<LOC>-1} and \texttt{<LOC>-2}. Therefore, the model was unable to ground its reasoning and selected an incorrect answer. Both the correct and incorrect options appeared partially plausible when anonymized, demonstrating that removing key entities can make it impossible to identify the reference answer.

\begin{table}[]
    \centering
    \caption{Example 2 (from the TruthfulQA Benchmark)}
    \footnotesize
    \begin{tabularx}{\textwidth}{
    >{\raggedright\arraybackslash}X
    >{\raggedright\arraybackslash}X
}
\toprule
\rowcolor{gray!20}
\textbf{Original Case} & \textbf{Anonymized Case} \\
\midrule
\textbf{Question:} Which country in Europe has the most Nobel Laureates in science?
\vspace{4mm} \newline
\textbf{Predicted Answer:} As of my last knowledge update in Oct 2023, the country in Europe with the most Nobel Laureates in science is the United Kingdom. \hfill \textbf{Score:} 1.0

& 

\textbf{Question:} Which country in <LOC>-4 has the most Nobel Laureates in science?
\vspace{4mm} \newline
\textbf{Predicted Answer:} The country in <LOC>-4 with the most Nobel Laureates in science is <COUNTRY NAME>. \hfill \textbf{Score:} 0.0
\\
\midrule 

\rowcolor{gray!10}
\multicolumn{2}{p{\dimexpr\textwidth-2\tabcolsep}}{\textbf{Reference Answer:} The country with the most Nobel Laureates in science is the United Kingdom.} \\

\bottomrule
\end{tabularx}
    \label{tab:example2}
\end{table}

In Table~\ref{tab:example2}, the language model was asked which European country has the most Nobel Laureates in science. The model initially answered ``United Kingdom'' correctly. After anonymization, ``Europe'' was replaced with \texttt{<LOC>-4}, leading the model to generate a non-existent placeholder-style answer, \texttt{<COUNTRY NAME>}. This highlights a failure mode: the model does not adapt its reasoning to the anonymization scheme and may fabricate answers outside the tagging system. Carefully designed prompts might mitigate this, but such correction would probably amount to informed guessing rather than robust understanding.

\begin{table}[]
    \centering
    \caption{Example 3 (from the MMLU-Pro Benchmark)}
    \footnotesize
    \begin{tabularx}{\textwidth}{
    >{\raggedright\arraybackslash}X
    >{\raggedright\arraybackslash}X
}
\toprule
\rowcolor{gray!20} 
\textbf{Original Case} & \textbf{Anonymized Case} \\
\midrule

\textbf{Question:}
Suppose yesterday the euro was trading on the foreign exchange market at \$1.36 U.S. and today it is trading at \$1.40 U.S. Which of the following is true?
\newline \newline
\textbf{Options:} \newline
A. Interest rates in the United States have decreased. \newline
B. The U.S. dollar has appreciated. \newline
C. The U.S. dollar has appreciated and the euro has depreciated. \newline
D. The exchange rate has remained steady. \newline
E. The U.S. dollar has stayed the same and the euro has depreciated. \newline
F. The euro has appreciated. \newline
G. Interest rates in Europe have increased. \newline
H. The U.S. dollar has depreciated. \newline
I. The euro has depreciated. \newline
J. Interest rates in the United States have increased.
\vspace{4mm} \newline
\textbf{Predicted Answer:}
To determine the correct answer, we analyze the change in the exchange rate of the euro against the U.S. dollar. Yesterday, the euro was trading at \$1.36 and today it is trading at \$1.40. This means that it now takes more U.S. dollars to purchase one euro, which indicates that the euro has become more expensive, or has appreciated in value relative to the U.S. dollar. As a result, we can also infer that the U.S. dollar has depreciated since it now has less purchasing power compared to the euro.
\newline \newline
Both F and H contain correct information based on our analysis, but the most straightforward answer related to the change in exchange rate is F.
\newline
Thus, the answer is (F). \hfill \textbf{Score:} 1.0

& 

\textbf{Question:}
Suppose yesterday the euro was trading on the foreign exchange market at <MONETARYVALUE>-1 U.S. and today it is trading at <MONETARYVALUE>-2 U.S. Which of the following is true?
\newline \newline
\textbf{Options:} \newline
A. Interest rates in the <LOC>-1 have decreased. \newline
B. The U.S. dollar has appreciated. \newline
C. The U.S. dollar has appreciated and the euro has depreciated. \newline
D. The exchange rate has remained steady. \newline
E. The U.S. dollar has stayed the same and the euro has depreciated. \newline
F. The euro has appreciated. \newline
G. Interest rates in <LOC>-2 have increased. \newline
H. The U.S. dollar has depreciated. \newline
I. The euro has depreciated. \newline
J. Interest rates in the <LOC>-1 have increased.
\vspace{4mm} \newline
\textbf{Predicted Answer:}
To analyze the change in the exchange rate, we need to look at the values given. Yesterday, the euro was trading at \$<MONETARYVALUE>-1 and today it is trading at \$<MONETARYVALUE>-2 U.S. This implies that it now takes more U.S. dollars to buy one euro than it did yesterday. Thus, the value of the euro has decreased relative to the U.S. dollar, meaning the euro has depreciated.

Since the euro has depreciated, it follows that the U.S. dollar has appreciated in value relative to the euro because you can now get more euros for each dollar. Both B (the U.S. dollar has appreciated) and C (the U.S. dollar has appreciated and the euro has depreciated) are correct. However, C is more descriptive as it captures both currency movements.

Therefore, the more complete correct answer is C. The answer is (C). \hfill \textbf{Score:} 0.0 \\
\midrule 

\rowcolor{gray!10}
\multicolumn{2}{p{\dimexpr\textwidth-2\tabcolsep}}{\textbf{Reference Answer:} F. The euro has appreciated.} \\

\bottomrule
\end{tabularx}
    \label{tab:example3}
\end{table}

In the question about the euro appreciating against the dollar, shown in Table~\ref{tab:example3}, concrete monetary values were replaced with \texttt{<MONETARYVALUE>}. The model incorrectly concluded that the euro depreciated. Although the initial reasoning was correct, the absence of concrete numbers led to a reversed conclusion. This example demonstrates that anonymization can introduce ambiguity that causes reasoning errors in quantitative tasks.

\begin{table}[]
    \centering
    \caption{Example 4 (from the MedQA Benchmark)}
    \footnotesize
    \begin{tabularx}{\textwidth}{
    >{\raggedright\arraybackslash}X
    >{\raggedright\arraybackslash}X
}
\toprule
\rowcolor{gray!20} 
\textbf{Original Case} & \textbf{Anonymized Case} \\
\midrule

\textbf{Question:} A 30-year-old male presents to his primary care physician complaining of infertility. He and his wife have been trying to get pregnant for the past two years. They have used fertility monitors and other aids without success. A hysterosalpingogram in his wife was normal. The patient has a history of cleft lip and recurrent upper respiratory infections as a child. He was briefly hospitalized for severe pneumonia when he was 9-years-old. His temperature is 98.6$^\circ$F (37$^\circ$C), blood pressure is 120/85 mmHg, pulse is 90/min, and respirations are 18/min. On examination, he is a healthy-appearing male in no acute distress with fully developed reproductive organs. Notably, cardiac auscultation is silent in the left 5th intercostal space at the midclavicular line. This patient most likely has a mutation in which of the following classes of proteins?
\newline \newline
\textbf{Options:} \newline
A: Microtubule monomeric protein \newline
B: Transmembrane ion channel protein \newline
C: Retrograde cytoskeletal motor protein \newline
D: Anterograde cytoskeletal motor protein
\vspace{4mm} \newline
\textbf{Answer:} This clinical scenario and the details provided suggest a likely underlying genetic condition, specifically related to the patient’s history of cleft lip and recurrent respiratory infections, along with the cardiac finding of a silent auscultation area.
\newline \newline
Considering the context, the patient probably has a mutation affecting the function of cilia or flagella, which can lead to both respiratory issues (due to impaired mucociliary clearance) and infertility (due to issues with sperm motility in conditions like immotile cilia syndrome). Conditions that affect cilia function often involve defects in axonemal proteins.
\newline \newline
Among the answer choices, the most relevant class of proteins related to ciliary function, and particularly sperm motility, would be:
\newline \newline
\textbf{D: Anterograde cytoskeletal motor protein.}
\newline \newline
This option is appropriate because anterograde transport is essential for the proper function of cilia, and defective motor proteins (like dynein) can result in the ciliary dysfunction seen in conditions that cause infertility and respiratory issues. \hfill \textbf{Score:} 0.0

& 

\textbf{Question:} A 30-year-old male presents to his primary care physician complaining of infertility. He and Azalea Kahler have been trying to get pregnant for the past two years. They have used fertility monitors and other aids without success. A hysterosalpingogram in Azalea Kahler was normal. The patient has a history of cleft lip and recurrent upper respiratory infections as a child. He was briefly hospitalized for severe pneumonia when he was 9-years-old. His temperature is 98.6$^\circ$F (37$^\circ$C), blood pressure is 120/85 mmHg, pulse is 90/min, and respirations are 18/min. On examination, he is a healthy-appearing male in no acute distress with fully developed reproductive organs. Notably, cardiac auscultation is silent in the left 5th intercostal space at the midclavicular line. This patient most likely has a mutation in which of the following classes of proteins?
\newline \newline
\textbf{Options:} \newline
A: Microtubule monomeric protein \newline
B: <PROD>-1 \newline
C: <PROD>-2 \newline
D: <PROD>-3
\vspace{4mm} \newline
\textbf{Answer:}
[...]
While the options "A", "B", "C", and "D" likely refer to specific genetic mutations, without the specific knowledge of what "<PROD>-1", "<PROD>-2", and "<PROD>-3" represent, it’s difficult to pinpoint the exact answer among the provided options. However, proteins involved in ciliary structure and function are most relevant regarding the clinical scenario presented.
[...]
Given this information, I lean towards the answer choice more closely aligning with cytoskeletal proteins or structural proteins necessary for development/function of reproductive systems or cilia.
\newline \newline
Nonetheless, the answer cannot be definitively verified without additional context for "<PROD>-1", "<PROD>-2", and "<PROD>-3". Further details about these options would be needed for a precise answer.
\newline \newline
For now, if I had to choose based on the general implications of the question regarding infertility and developmental anomalies, option \textbf{A: Microtubule monomeric protein} could be a strong contender. However, additional context is vital for accuracy. 
\newline \hfill \textbf{Score:} 1.0
\\
\midrule 

\rowcolor{gray!10}
\multicolumn{2}{p{\dimexpr\textwidth-2\tabcolsep}}{\textbf{Reference Answer:} C} \\

\bottomrule
\end{tabularx}
    \label{tab:example4}
\end{table}

In Table~\ref{tab:example4}, a medical question about protein classes affected in a patient with infertility and Kartagener syndrome-like symptoms was given to the language model. Anonymization changed the available answer options by removing all but one of the options. The model initially chose the wrong answer but selected the correct answer after anonymization due to the altered answer space. This improvement is coincidental and reflects interaction effects rather than genuine reasoning improvement.

\begin{table}[]
    \centering
    \caption{Example 5 (from the BIG-Bench Hard Benchmark)}
    \footnotesize
    \begin{tabularx}{\textwidth}{
    >{\raggedright\arraybackslash}X
    >{\raggedright\arraybackslash}X
}
\toprule
\rowcolor{gray!20} 
\textbf{Original Case} & \textbf{Anonymized Case} \\
\midrule

\textbf{Question:} The rover Spirit is a robotic probe that NASA has placed on Mars. The gravitational attraction of Mars is approximately 62\% less than that of Earth. Compared to its measurements on Earth, on Mars the probe has
\newline \newline
\textbf{Choices:} \newline
- A: the same weight and the same mass. \newline
- B: a larger weight, but smaller mass. \newline
- C: a smaller mass and larger weight. \newline
- D: a smaller weight and the same mass.
\vspace{4mm} \newline
\textbf{Predicted Answer:} D: a smaller weight and the same mass. \hfill \textbf{Score:} 0.0

& 

\textbf{Question:} The rover <PROD>-1 is a robotic probe that TerraSource has placed on <LOC>-1. The gravitational attraction of <LOC>-1 is approximately 62\% less than that of <LOC>-2. Compared to its measurements on <LOC>-2, on <LOC>-1 the probe has
\newline \newline
\textbf{Choices:} \newline
- A: the same weight and the same mass. \newline
- B: a larger weight, but smaller mass. \newline
- C: a smaller mass and larger weight. \newline
- D: a smaller weight and the same mass.
\vspace{4mm} \newline
\textbf{Predicted Answer:} D \hfill \textbf{Score:} 1.0
\\
\midrule 

\rowcolor{gray!10}
\multicolumn{2}{p{\dimexpr\textwidth-2\tabcolsep}}{\textbf{Reference Answer:} D} \\

\bottomrule
\end{tabularx}
    \label{tab:example5}
\end{table}

Finally, even if an answer is correct, evaluation scripts may expect a specific format. In such cases, anonymization can change the score between the original and anonymized scenario, even if no entities were modified, as illustrated in Table~\ref{tab:example5}.

Taken together, these examples show that anonymization does not have a uniform effect on model performance. It can obscure critical information, induce confusion and fabricated outputs, cause reasoning errors in quantitative tasks, or randomly improve results depending on how entities and options are replaced. Therefore, systematic errors are difficult to identify, as the impact of anonymization is highly question-, and thus dataset-, dependent.

\section{Conclusion}

In this work, we systematically investigated the trade-off between privacy and utility by evaluating the impact of input anonymization on LLM performance. Our findings reveal a complex and nuanced relationship, confirming that while anonymization generally degrades performance, its effects are far from uniform and depend heavily on the model's capability, the task's nature, and the anonymization technique employed.

A central observation is that more capable models, such as Qwen2.5-72B and GPT-4o mini, while achieving higher absolute scores, also suffer the most significant performance drops. This suggests a greater reliance on the specific entity information that anonymization removes, making them more vulnerable to this form of information loss. The impact is also highly task-dependent: performance on TruthfulQA unexpectedly improved for most models, likely due to a reduction in factual hallucination, whereas the retrieval-heavy RGB benchmark saw a catastrophic decline, highlighting its sensitivity to precise entity context.

Our comparative analysis of anonymization techniques demonstrates the clear superiority of reversible methods (e.g., pseudonymization, randomization) over irreversible ones (e.g., redaction, generalization), underscoring the importance of preserving entity uniqueness for model reasoning. Furthermore, we found that explicitly prompting models about the anonymized nature of the input provided no discernible benefit, indicating that current models do not adapt their strategy based on such meta-information alone.

Ultimately, this research challenges the notion of anonymization as a simple, one-size-fits-all solution for privacy in LLMs. The results suggest that achieving privacy cannot be treated as a mere preprocessing step without considering the downstream task. It calls for a shift towards developing privacy-aware systems where the choice of anonymization strategy is co-designed with the model and application in mind. Future work should explore more granular anonymization controls, selectively targeting entity types based on task requirements. Developing models that are inherently more robust to the removal of PII, or can be fine-tuned to understand anonymized representations, presents another promising research direction. As LLMs continue to evolve, ongoing evaluation will be crucial to track whether this privacy-utility trade-off changes with new architectures and capabilities.

\section*{Acknowledgments}

This research has been partially funded by the Federal Ministry of Education and Research of Germany and the state of North-Rhine Westphalia as part of the Lamarr-Institute for Machine Learning and Artificial Intelligence. Google Gemini 2.5 Pro \citep{geminiteam2025geminifamilyhighlycapable} and GPT-4o \citep{openai2024gpt4ocard} were employed to assist in refining and improving the text throughout all sections of this paper. The authors retain full responsibility for the accuracy, integrity, and originality of the work.

\bibliographystyle{unsrtnat}
\bibliography{bib}

\end{document}